\pdfoutput=1

\documentclass[11pt]{article}

\usepackage[final]{acl}

\usepackage{times}
\usepackage{latexsym}

\usepackage[T1]{fontenc}

\usepackage[utf8]{inputenc}

\usepackage{microtype}

\usepackage{inconsolata}

\usepackage{graphicx}
\usepackage{amsmath}
\usepackage{algorithm}
\usepackage{algorithmic}

\title{Data Descriptions from Large Language Models with Influence Estimation}

\author{Chaeri Kim\thanks{These authors contributed equally to this work.} \qquad  Jaeyeon Bae\footnotemark[1] \qquad Taehwan Kim \\ 
  Ulsan National Institute of Science and Technology(UNIST) \\
  \{chaerikim, qowodussla, taehwankim\}@unist.ac.kr 
}

\begin{document}

\maketitle
\begin{abstract}
Deep learning models have been successful in many areas but understanding their behaviors still remains a black-box. Most prior explainable AI (XAI) approaches have focused on interpreting and explaining how models make predictions. In contrast, we would like to understand how data can be explained with deep learning model training and propose a novel approach to understand the data via one of the most common media - language - so that humans can easily understand. Our approach proposes a pipeline to generate textual descriptions that can explain the data with large language models by incorporating external knowledge bases. However, generated data descriptions may still include irrelevant information, so we introduce to exploit influence estimation to choose the most informative textual descriptions, along with the CLIP score. Furthermore, based on the phenomenon of cross-modal transferability, we propose a novel benchmark task named ~\emph{cross-modal transfer classification} to examine the effectiveness of our textual descriptions.  In the experiment of zero-shot setting, we show that our textual descriptions are more effective than other baseline descriptions, and furthermore, we successfully boost the performance of the model trained only on images across all nine image classification datasets. These results are further supported by evaluation using GPT-4o. Through our approach, we may gain insights into the inherent interpretability of the decision-making process of the model.

\end{abstract}

\section{Introduction}
\label{sec:intro}

Deep learning models have successfully been applied to various fields and achieved high performance~\citep{huang2017densely, vaswani2017attention, he2016deep, dosovitskiy2020image}. Despite the rapid performance improvement, understanding their behaviors remains a black-box. While most prior explainable AI (XAI) approaches have focused on interpreting and explaining how models make predictions, there are few attempts to explain the data. We would like to understand data since it is one of the most important elements of deep learning performance. Among deep learning models, we try to describe image classes with \emph{human-interpretable} language. 
For example, in the case of frog class: smooth, moist skin with coloration ranging from green to brown, often featuring various patterns and markings.

Recent studies~\citep{menon2022visual, maniparambil2023enhancing, pratt2023does} have utilized large language models (LLMs), pre-trained on vast datasets, to generate textual descriptions that aid vision model performance. However, identifying descriptions that are truly informative and semantically aligned with the class remains a key challenge.
To solve this problem, we propose a novel approach to identify the most helpful texts for model training among those generated by the LLM, using influence estimation and CLIP scores. Influence estimation \cite{koh2017understanding, pruthi2020estimating} calculates the positive or negative impact of each training image in predicting a test sample, while CLIP scores measure the similarity between an image and text.
Unlike most previous research that applies influence scores only to images, we extend this method to texts by integrating CLIP scores to identify helpful texts for model training. We name the helpful text as \emph{proponent text} and the score for determining the proponent text as \emph{Influence scores For Texts} (IFT).

\begin{figure*}[!h]
\centering
\includegraphics[width=0.95\textwidth]{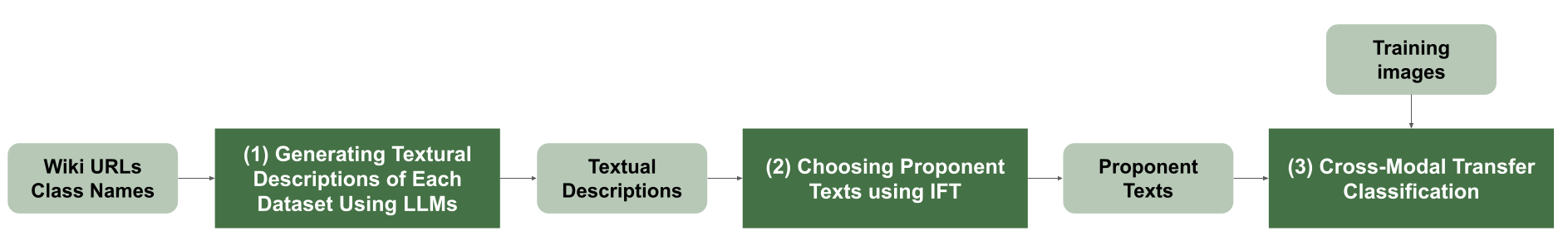}
\caption{
Overview of our framework. (1) 'Generating Textual Descriptions of Each Dataset' process generates the textual descriptions that can explain each class well from given Wikipedia urls, pre-defined prompts, and class names
(2) 'Choosing Proponent Texts using IFT' process determines the proponent texts using IFT.
(3) 'Cross-Modal Transfer Classification' process first trains model with images, then further trains it with proponent texts.
}
\label{fig:overview}
\vspace{-0.2cm}
\end{figure*}

To obtain the proponent texts, we first write down all the features of each image class with GPT-3.5~\citep{brown2020language}. Instead of merely querying the LLM for class features, we employ a two-stage prompting process using Wikipedia urls to integrate external knowledge bases.
This approach not only compensates for potential gaps in detailed knowledge of LLM but also helps mitigate hallucinations. Among the generated textual descriptions, IFT filters out unhelpful information and retains only proponent texts. 
Furthermore, leveraging the concept of cross-modal transferability~\cite {zhangdrml}, which allows text to serve as input for models originally trained with images, we apply this principle to our approach. 
We retrain the image-trained model using proponent texts, assigning weights based on the IFT. We call this \emph{'cross-modal transfer classification'}.
Compared to image-only training and baselines that generate textual descriptions, our approach achieves superior performance on most of the nine image datasets, demonstrating that our textual descriptions are more helpful in model training than other baselines. Additionally, we evaluate the helpfulness, informativeness, and relevance of the generated descriptions using GPT-4o~\cite{hurst2024gpt} and find that our method consistently outperforms all baselines across these criteria. Figure \ref{fig:overview} shows the overview of our framework.

In sum, our contributions are as follows:
\begin{itemize}
\item We propose a novel approach to identify helpful textual descriptions that effectively explain each image class, achieving superior performance in the zero-shot setting compared to baseline descriptions.
\item By using the proposed IFT defined as the sum of the influence score and the CLIP score, we can pinpoint the most helpful textual descriptions for model training. Proponent texts selected through IFT contain only relevant information that aids in accurate image classification. 
Furthermore, we may get an inherent interpretability of where the black-box model focuses on during training.

\item We propose \emph{cross-modal transfer classification} as a novel benchmark task based on the cross-modal transferability phenomenon. Extensive experiments with nine datasets show that the proponent texts are informative and describe each image class well. 
\end{itemize}

\section{Related Work}
\label{sec:related}

\begin{figure*}
\centering
\includegraphics[width=\textwidth]{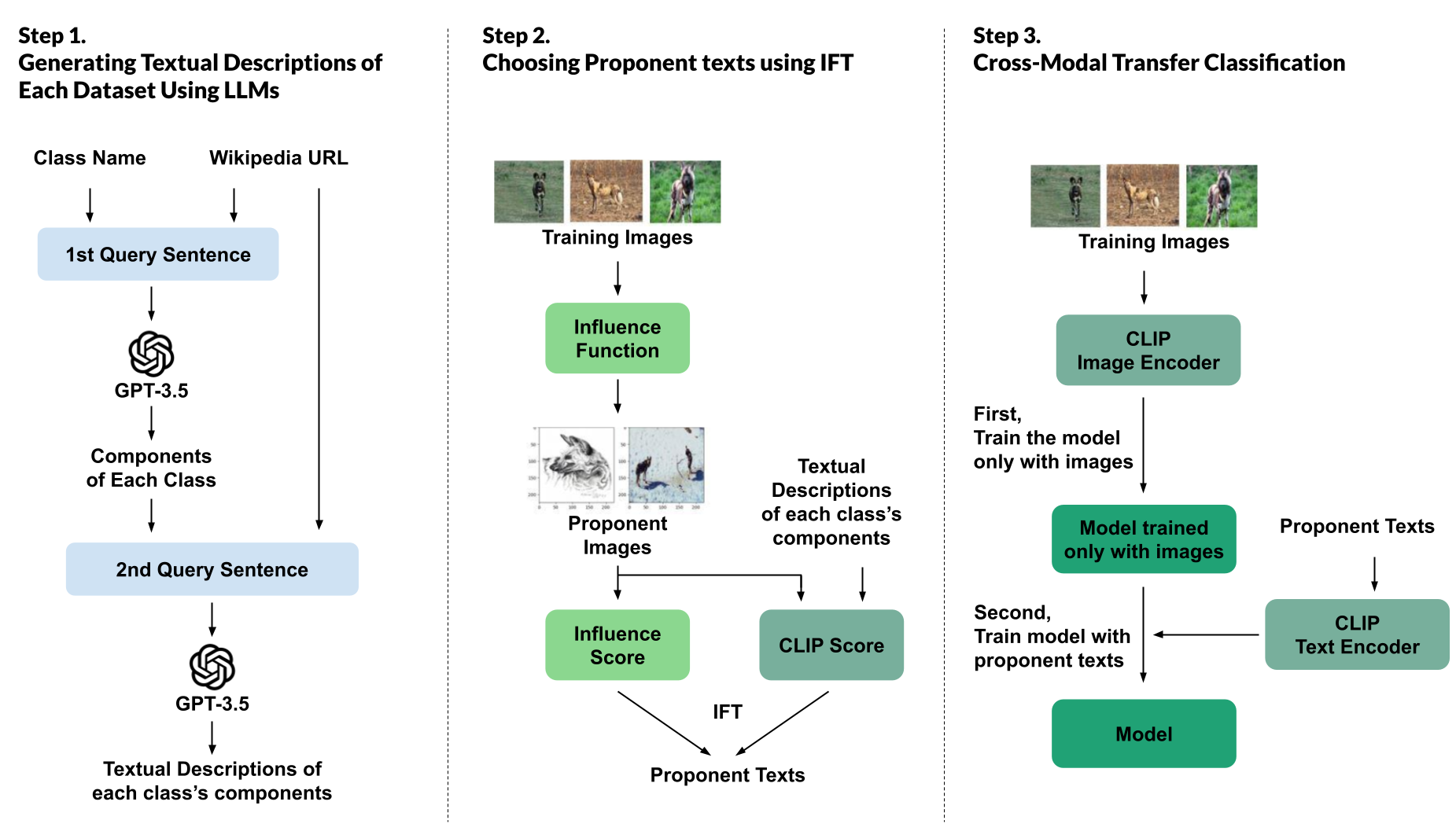}
\caption{
Details of our approach. 
(1) Extract class components from class names and obtain textual descriptions using Wikipedia urls.
(2) Identify proponent images using influence scores, then combine CLIP scores and influence scores to get proponent texts
(3) Train the model with training images, followed by cross-modal transfer training using proponent texts.
}
\label{fig:Approaches}
\vspace{-0.2cm}
\end{figure*}

 Not only achieving high accuracy, but most research has focused on why the model makes such decisions. Influence function is one of the powerful tools for explaining model decisions. One widely used approach~\citep{koh2017understanding} examines changes in model parameters when input data is perturbed, using second-order optimization techniques to efficiently approximate influence scores. TracIn~\citep{pruthi2020estimating} calculates the influence of training data on the loss of a test sample by tracing how loss changes during the training process with a fixed test sample.

Efforts to enhance model explainability have also extended to the use of large language models (LLMs). Language Guided Bottlenecks (LaBo)~\citep{yang2023language} is an extended model of Concept Bottleneck Models~\citep{koh2020concept} that queries LLMs to collect concepts. \citet{menon2022visual} uses descriptions generated through LLMs and proposes an alternative zero-shot classification method named 'classification by description'. However, \citet{menon2022visual} highlights some limitations: their generated text descriptions contain non-useful visual cues for the vision-language model and sometimes include repeated text. VDT-Adapter~\cite{maniparambil2023enhancing} and CuPL~\cite{pratt2023does} two approaches that also leverage LLMs to enhance vision-language models. VDT-Adapter~\cite{maniparambil2023enhancing} improves the image classification performance of CLIP by using GPT-4 to generate visually descriptive text prompts, helping the model focus on relevant visual details. Similarly, CuPL~\cite{pratt2023does} (Customized Prompts via Language models) generates category-specific prompts using LLMs, improving zero-shot classification without additional training, and outperforming hand-crafted prompts across multiple benchmarks. Nevertheless, both approaches have limitations in identifying the most relevant textual descriptions.

To deal with these issues, we define \emph{influence scores for texts} (IFT) composed of influence score and CLIP score. IFT serves as a scoring metric to determine the importance of each text. Through the proposed IFT, our proponent texts include only the information necessary for precise image classification.
Moreover, the advent of models trained with vision-language contrastive learning, such as CLIP~\citep{radford2021learning}, provides cross-modal transferability phenomenon~\citep{zhangdrml}. By integrating these insights, we propose a novel benchmark task \emph{cross-modal transfer classification}, which underscores the effectiveness of our textual descriptions in enhancing model performance.

\section{Approach}
\label{sec:approach}

\begin{figure*}[h]
\begin{align}
    \text{Influence Score}(i^{t}, i^{v}) = \sum_{j=1}^{k} \eta _{j} \nabla loss(w_{t_{j}}, i^{t}) \cdot \nabla loss(w_{t_{j}}, i^{v}))\quad i^{t} \in I^{train}, i^{v} \in I^{val} \label{eq:IF}\ \\
    \text{IFT}(T_{c'})= \frac{1}{n_{I^{val}} \cdot n_{I^{train}}} \sum_{i^v \in I^{val}} \sum _{i^t \in I^{train}} (\text{Influence Score}(i^{t}, i^{v}) + \text{CLIP Score}(i^{t}, T_{c'}))) \label{eq:IFT}
\end{align}
\vspace{-0.4cm}
\end{figure*}
 
As described in Figure \ref{fig:Approaches}, our framework consists of three main steps. In this section, we describe each step in detail.

\subsection{Generating Textual Descriptions of Each Dataset Using LLMs} 
\label{sec:subapproach1}

\textbf{Extracting Components of Each Class} From the raw class names of the dataset, we generate questions by placing the class names into a pre-defined prompt. For datasets with subcategories under a superclass, we query GPT-3.5 for the components of the superclass. For example, in the CUB 200 2011~\citep{wah2011caltech} dataset, "bird" is the superclass, and species like "Laysan Albatross" and "Fish Crow" are subcategories. In this case, we ask about the components of "birds" rather than the individual species.
The question format is as follows:
"\emph{Q : Can you tell me the components of $\{$class name$\}$ from the perspective of appearance? A : }". 
The extracted components are then used to extract the textual descriptions of each image class. For example, for the 'African hunting dog' class in the Miniimagenet dataset~\citep{vinyals2016matching}, the components include: body build, coat color, ears, head, and eyes.

\textbf{Generating Textual Descriptions of Each Class} Knowing the components of each class, we can query GPT-3.5 about the appearance of each component. 
To avoid lacking detailed knowledge and hallucinations, we provide the corresponding Wikipedia url. 
GPT-3.5 summarizes the relevant information from the Wikipedia url in one line, composed of nouns. If no relevant information is found on Wikipedia, we ask GPT-3.5 to provide a summary based on its existing knowledge. The question format is as follows:
"\emph{Q : Please summarize the information of appearance about $\{$components of each class$\}$ in this $\{$Wikipedia url$\}$ in one line composed of nouns. If you couldn't find related information, you must answer general information you know. A : }"

\subsection{Choosing Proponent Texts} 
\label{sec:subapproach2}

We use TracIn~\citep{pruthi2020estimating} to calculate influence scores for images. Based on influence scores, we identify proponent images from training images that aid in predicting each validation image.
Then, we calculate the CLIP score, which refers to the cosine similarity between the CLIP embeddings of the proponent images and the corresponding textual descriptions in the validation set.
Higher CLIP scores indicate stronger semantic alignment between images and descriptions.
With the sum of the influence score and CLIP score, we can calculate \emph{influence scores for texts} (IFT). For notations, let $I^{train}$ and $I^{val}$ denote image samples of the train and validation dataset respectively and $T_C$ denotes the textual descriptions for all classes $\mathcal{C}$. $n_{I^{train}}$ is the number of image samples in the training dataset and $T_{c}$ are the extracted textual descriptions of class $c$ where $c \in \mathcal{C}$.

In equation \eqref{eq:IF}, the loss of model parameterized by $w$ on training image sample $i^t$ can be denoted as $loss(w, i^t)$. 
In this context, $\Delta$loss specifically refers to the cross-entropy loss, as the task involves image classification.
Influence score measures the impact of specific training examples on a given validation sample. Since considering only one training sample at a time is impractical, TracIn~\citep{pruthi2020estimating} introduces practical influence function via \textit{k} checkpoints ${w_{t_{1}}, w_{t_{2}}, ..., w_{t_{k}}}$ and minibatches through simple first-order approximation. 
It is the total reduction in loss on a fixed validation example $i^{v}$ in the training process. $\text{CLIP score} (i^{t}, T_{c})$ computes the correlation between CLIP image embeddings of proponent training image $i^{t}$ and CLIP text embeddings of textual description for class $c$ $T_{c}$. 

Equation \eqref{eq:IFT} defines the IFT score. We calculate the average of the sum of influence scores and CLIP scores for the textual descriptions of each image class. We then select ten textual descriptions with the highest IFT scores, naming them \emph{proponent texts}. These proponent texts help us effectively explain and understand each image class.

\begin{algorithm*}[h]
    \caption{Cross-Modal Transfer Classification Training Algorithm} 
    \label{alg:cmtm}
    \scalebox{0.9}{
    \begin{minipage}{\linewidth}
    \begin{algorithmic}[1]
        \STATE \textbf{Input:} 
        \STATE \quad Training images : $I^{train}$
        \STATE \quad Textual description for class c : ${T_c}$
        \STATE \quad Frozen CLIP encoders : $\text{CLIP}_{\text{ImageEncoder}}, \text{CLIP}_{\text{TextEncoder}}$
        \STATE \textbf{Output:} classifier $M$
        \STATE \quad $e_{i_t} \gets \text{CLIP}_{\text{ImageEncoder}}(i_t)$ for each $i_t\in I^{train}$
        \STATE \quad $e_{T_c} \gets \text{CLIP}_{\text{TextEncoder}}(T_c)$ for each $c\in\mathcal{C}$
        \STATE \quad $w_c \leftarrow \mathrm{IFT}_c \big/ \sum_{c \in \mathcal{C}}\mathrm{IFT}_{c}$
        \STATE \quad \textbf{Step 1 (image):} Train $M$ with $\mathcal{L}_{img} = CE(M(e_{i_t}), L_{i_t})$
        \STATE \quad \textbf{Step 2 (text):} Train $M$ with $\mathcal{L}_{txt} = CE(M(w_c \cdot e_{T_c}), L_{T_c})$
        \RETURN $M$
    \end{algorithmic}
    \end{minipage}}
\end{algorithm*}

\subsection{Cross-Modal Transfer Classification} 
\label{sec:subapproach3}

Cross-modal transferability~\citep{zhangdrml} states that text inputs can work as good proxies to image inputs trained on a shared image-text embedding space obtained through multi-modal learning. Based on this phenomenon, we can use texts as inputs instead of images for vision models trained with images if images and texts are in a shared embedding space.

We use CLIP~\citep{radford2021learning} image encoder and text encoder to align images and proponent texts in the same embedding space. Denote the dimension of CLIP embedding space as $D$. An input training image $i^t$ is projected into the image embedding space as $e_{i^t} = CLIP_{ImageEncoder}(i^t)$, where $e_{i^t} \in \mathcal{R}^{D}$. Similarly, proponent text $T_{c}$ for class $c$ is projected into a text embedding $e_{T_{c}} = CLIP_{TextEncoder}(e_{c})$ where $e_{t_{c}} \in \mathcal{R}^{D}$.
The model, which consists of linear layers, is initially trained on image embeddings and subsequently retrained on proponent text embeddings. While training the model with proponent texts $T_\mathcal{C}$, weights are given for each $e_{T_{c}}$ based on their normalized IFT scores : $\sum _{c \in \mathcal{C}} w_c \cdot CELoss($\( \hat{y_{T_c}} \)$, y_{T_c})$ while $w_c = \frac{IFT_c}{\sum_{c \in \mathcal{C}} IFT_{c}}$. Here, $CELoss$ refers to the cross-entropy loss, which measures the discrepancy between the predicted labels \( \hat{y_{T_c}} \) and the true labels $y_{T_c}$ of the classes. Because $e_{i^t}$ and $e_{T_{c}}$ share the same embedding space $\mathcal{R}^{D}$, the model can be trained without any additional implementations. 
For clarity, we include the complete training procedure in Algorithm~\ref{alg:cmtm}, which first train the classifier only with the training images and then refines it with weighted proponent texts. By combining these two complementary training steps, our approach leverages both modalities in a unified embedding space, thereby enabling effective cross-modal transfer classification.

This method is simple yet effective, as it improves performance by only updating the linear layer while the CLIP encoder remains frozen. Furthermore, our approach has a low computational cost. Training with only images on all datasets takes about 2 hours using an NVIDIA 3090 GPU, while cross-modal transfer training with proponent texts requires less than 30 minutes. 
Additionally, the performance improvement indicates that the proponent texts effectively explain each image class and enhance the training process. 

Our novel \emph{cross-modal transfer classification} benchmark aims to improve image classification model performance using effective and helpful textual descriptions. We expect this benchmark to enhance understanding of black-box models by improving performance through human-readable language. Furthermore, it may provide an inherent explanation of the decision-making process of the black-box model.

\begin{table*}[h]
\centering
\resizebox{0.9\textwidth}{!}{%
\begin{tabular}{c|c|c|c|c|c|c}
\hline
{Datasets} & {CLIP Zero-shot} & {Menon} & {LABO} & {CuPL} & {VDT-Adapter} & {Ours Zero-Shot} \\ \hline
{CUB 200 2011} & {38.540\%} & {52.969\%} & {52.917\%} & {\textbf{53.349\%}} & {53.162\%} & {{\underline{53.227\%}}} \\
{OxfordPets} & {81.132\%} & {85.580\%} & {87.196\%} & {\textbf{88.814\%}} & {88.140\%} & {{\underline{88.679\%}}} \\
{CIFAR10} & {88.800\%} & {\underline{89.320\%}} & {88.709\%} & {88.150\%} & {89.090\%} & {\textbf{89.470\%}} \\
{CIFAR100} & {61.680\%} & {\underline{63.999\%}} & {60.460\%} & {63.660\%} & {63.450\%} & {\textbf{64.570\%}} \\
{EuroSat} & {30.815\%} & {32.630\%} & {28.667\%} & {\underline{38.925\%}} & {38.457\%} & {\textbf{39.148\%}} \\
{Food101} & {80.620\%} & {\textbf{83.644\%}} & {82.871\%} & {83.339\%} & {83.013\%} & {\underline{83.452\%}} \\
{Miniimagenet} & {81.630\%} & {84.780\%} & {84.890\%} & {84.720\%} & {\underline{85.199\%}} & {\textbf{85.320\%}} \\
{102flowers} & {58.730\%} & {66.670\%} & {66.667\%} & {67.643\%} & {\underline{68.742\%}} & {\textbf{69.109\%}} \\
{DTD} & {43.085\%} & {37.660\%} & {46.489\%} & {47.553\%} & {\underline{48.457\%}} & {\textbf{48.989\%}} \\ \hline
\end{tabular}
}
\caption{
\label{tab:zero_shot_result}
Accuracy for test images in zero-shot setting. The best performing ones in bold font and underlined represent
the second-best performance.
}
\vspace{-0.2cm}
\end{table*}
\section{Experiments}
\label{sec:exp}

In the following sections, we detail our experimental setup, present the results of our approach in comparison. Additionally, we perform ablation studies to further evaluate the contributions of individual components and their impact on overall performance.

\subsection{Experimental Setup and Details} \label{sec:expdetails}

\textbf{Datasets} We use nine image datasets for our experiments: CUB 200 2001~\citep{wah2011caltech}, Miniimagenet~\citep{vinyals2016matching}, CIFAR-10~\citep{krizhevsky2009learning}, CIFAR-100~\citep{krizhevsky2009learning}, OxfordPets~\citep{parkhi2012cats}, EuroSAT~\citep{helber2019eurosat}, Food101~\citep{bossard2014food}, 102flowers~\citep{nilsback2008automated}, and Describable Textures Dataset (DTD)~\citep{cimpoi2014describing}. 
We follow the official dataset partitions when available, and use 20\% of the training set as the validation set when no official validation set is provided. If no official train/test split exists, we randomly divide the dataset into training, validation, and test sets. 
Detailed dataset partitions are reported in Section \ref{apd:dataset} of Appendix. 

\textbf{Implementation Details} We use ViT/32 CLIP~\citep{radford2021learning} as the image encoder.
We train the linear model using stochastic gradient descent (SGD) with a mini-batch size of 64 with learning rate of 0.1. The vision model, initially trained only with images, is further trained with proponent texts for a total of 30 epochs in all settings. We also use the CosineAnnealingLR learning scheduler with the maximum number of iterations of 200.

\begin{table*}[h]
\centering
\resizebox{0.8\textwidth}{!}{%
\begin{tabular}{c|c|c|c|c|c|c}
\hline
Datasets & \begin{tabular}[c]{@{}c@{}}Only\\ Images\end{tabular} & \begin{tabular}[c]{@{}c@{}}Menon\end{tabular} & LaBo & CuPL & VDT-Adapter & Ours \\ \hline
CUB 200 2011 & 71.332\% & 74.905\% & 73.991\% & \underline{74.957\%} & 73.699\% & \textbf{75.130\%} \\
OxfordPets & 91.664\% & \underline{91.711\%} & 90.700\% & 91.105\% & 91.431\% & \textbf{93.396\%} \\
CIFAR10 & \underline{94.320\%} & 93.470\% & 93.480\% & 93.420\% & 93.520\% & \textbf{94.820\%} \\
CIFAR100 & 77.830\% & 78.380\% & 78.400\% & 78.270\% & \underline{78.430\%} & \textbf{78.650\%} \\
EuroSat & 94.296\% & 93.518\% & 95.259\% & \underline{95.293\%} & 94.482\% & \textbf{96.037\%} \\
Food101 & 85.848\% & \underline{86.402\%} & 86.165\% & 86.145\% & 86.138\% & \textbf{86.950\%} \\
Miniimagenet & 91.980\% &  \underline{92.030}\%&  91.770\%&  91.640\%&  91.740\%& \textbf{92.480\%} \\
102flowers & 96.459\% & 96.825\% & 96.948\% & 96.581\% & \underline{97.191\%} & \textbf{97.948\%} \\
DTD & 72.074\% & 72.872\% & 72.713\% & 72.446\% & \underline{72.767\%} & \textbf{74.393\%} \\ \hline
\end{tabular}
}
\caption{Accuracy for test images when training using only images (Only Images) and cross-modal transfer training with texts (Cross-Modal Transfer Classification). The best performing ones in bold font and underlined represent the second-best performance.}
\label{tab:cross_modal_result}
\vspace{-0.2cm}
\end{table*}

To calculate influence scores, we use checkpoints during training. We train a pre-trained ResNet34 with stochastic gradient descent, a mini-batch size of 64, and a learning rate starting at 0.1, divided by 10 every 30 epochs, terminating at 200 epochs. We use checkpoints every 10 epochs.

\textbf{Baselines} We compare our method with other approaches that use large language model to generate text descriptions for image classes~\citep{menon2022visual, maniparambil2023enhancing, pratt2023does, yang2023language}. 
For each baseline, We use the pre-generated descriptions available in the official repositories, adjusting them to fit our experimental setup. 
For datasets where descriptions are not provided, we reproduce the authors' code to generate the corresponding descriptions.


\subsection{Result} \label{sec:results}

\begin{figure*}
\centering
\includegraphics[width=0.9\textwidth]{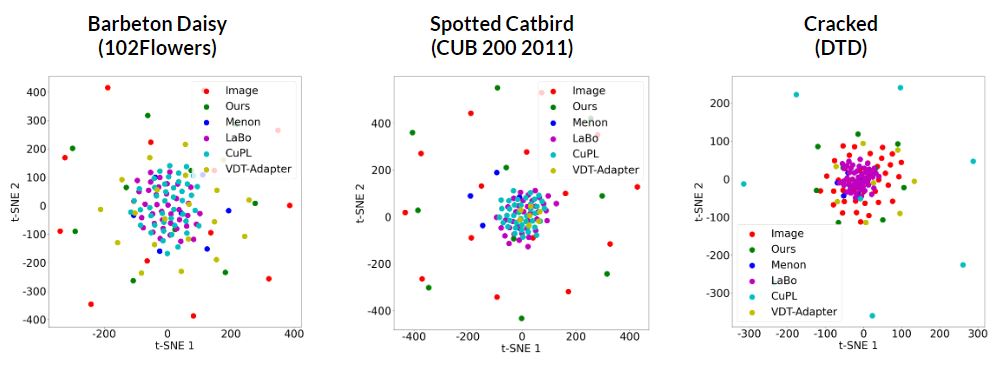}
\caption{Visualization of the embeddings of textual descriptions of our method and baseline method and the embeddings of the same class images with t-SNE. (a) Barbeton Daisy class of 102flowers dataset (b) Spotted catbird class of CUB 200 2011 dataset (c) cracked class of DTD dataset}
\label{fig:tsne}
\vspace{-0.2cm}
\end{figure*}

Table \ref{tab:zero_shot_result} reports the performance of our approach in zero-shot setting, comparing our textual descriptions with the baseline methods. This setting aims to demonstrate the effectiveness of our textual descriptions in a zero-shot context, aligning with the zero-shot setting assumptions in Menon~\cite{menon2022visual}.
In our approach, we utilize GPT-3.5, and for a fair comparison, we also apply GPT-3.5 to methods that originally used earlier versions, while employing GPT-4 for methods that used GPT-4.
Our method outperforms the baselines in most datasets. This result demonstrates that our textual descriptions explain each image class better than the baseline method. Additionally, our 2-stage prompting with Wikipedia urls, proves to be effective, outperforming not only methods that use the same GPT-3.5 but also those that utilize the more advanced GPT-4. 
To verify the robustness of our method, we conduct additional zero-shot classification experiments using a newer vision-language model, laion/CLIP-ViT-L-14-laion2B-s32B-b82K \cite{cherti2023reproducible}. This model is trained on a larger dataset and reflects recent advances in CLIP training. As shown in Table~\ref{tab:appendix_clip_new} in the Appendix, our method maintains strong performance across diverse datasets, consistently outperforming baseline methods and demonstrating that our method remains effective across CLIP architectures and continues to outperform most baselines.



Furthermore, as shown in Table \ref{tab:cross_modal_result}, most approaches achieve higher performance with cross-modal transfer training compared to training with images alone. Notably, our approach exhibits the largest performance gains across all datasets when proponent texts are incorporated into cross-modal transfer training. This result further supports that our proponent texts can explain each image class well and the prediction of the black-box model.


As our method consistently outperforms other baselines in both the zero-shot setting and cross-modal transfer classification, this suggests that the improvements stem not from the mere inclusion of textual descriptions but from selecting the most informative ones to enhance model training. If these gains are due to the model’s inherent preference for textual supervision (i.e., inductive bias), similar improvements would be observed across all textual description baselines. However, as shown in Table \ref{tab:zero_shot_result} and Table \ref{tab:cross_modal_result}, our method consistently achieves greater improvements, indicating that performance gains arise from our influence-guided selection of proponent texts rather than the presence of textual descriptions alone.

\begin{figure}
\centering
\includegraphics[width=0.45\textwidth]{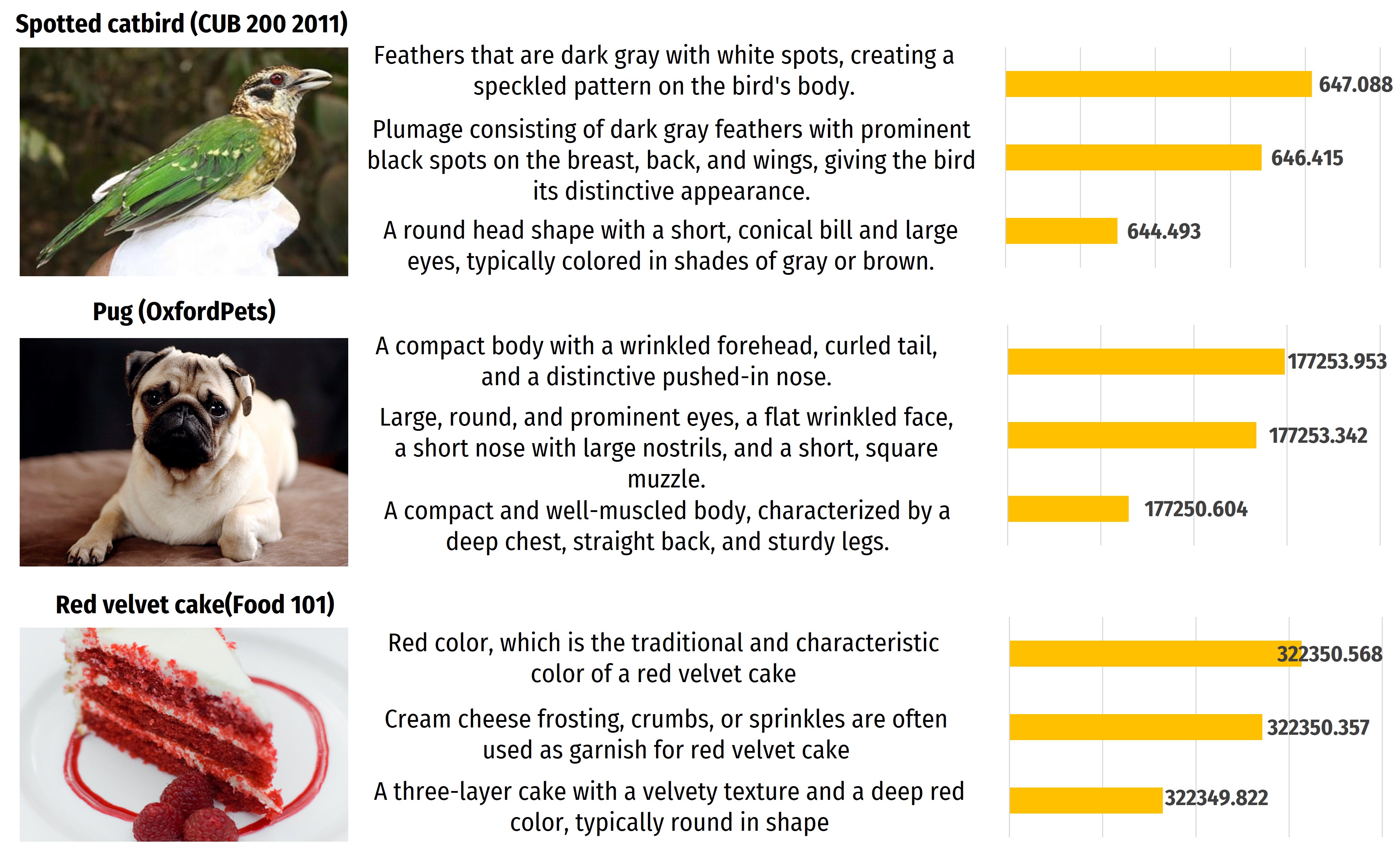}
\caption{Examples of proponent images, proponent texts and IFT for three classes. (Best viewed at an increased zoom level for clearer details.)}
\label{fig:prop_img_text_example} 
\vspace{-0.3cm}
\end{figure}

To investigate why our method performs better than the baselines, we visualize the embeddings of each textual description and the embeddings of the same class images with t-SNE~\citep{van2008visualizing}. 
Figure \ref{fig:tsne} presents the t-SNE visualizations of selected classes from the 102flowers~\cite{nilsback2008automated}, CUB 200 2011~\cite{wah2011caltech}, and DTD~\cite{cimpoi2014describing} datasets, comparing the image embeddings with text embeddings from our method and various baselines. In all three datasets, our method appears to align more closely with the image embeddings compared to the baselines. For instance, in the CUB 200 2011 dataset, the Spotted Catbird class shows that while CuPL~\cite{pratt2023does}, LaBo~\cite{yang2023language}, and VDT-Adapter~\cite{maniparambil2023enhancing} tend to cluster primarily in the center, our method achieves better alignment with the corresponding image embeddings. These results demonstrate that our approach leads to superior alignment between images and textual descriptions, thereby improving performance in cross-modal transfer tasks.
Furthermore, Figure \ref{fig:prop_img_text_example} provides examples of proponent images and proponent texts and analysis in Section \ref{apd:analysis} of Appendix demonstrates how IFT effectively selects texts that explain each image class by comparing proponent texts with non-proponent texts.

\begin{figure}
\centering
\includegraphics[width=8cm]{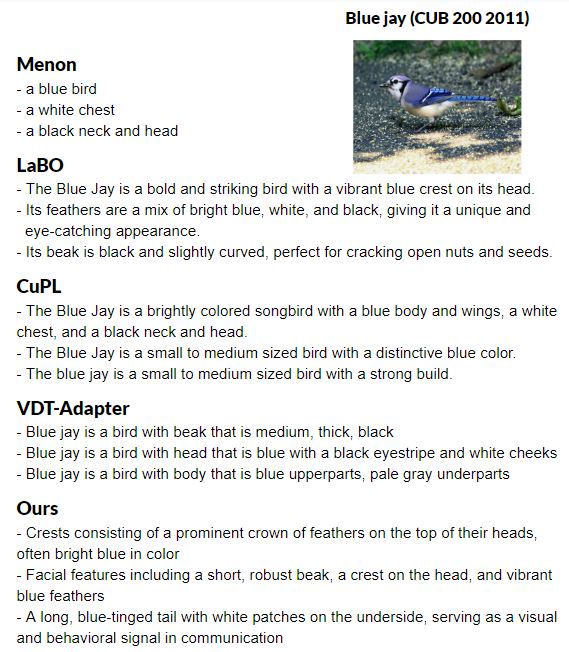}
\caption{\label{fig:human_eval_main} Example of a qualitative sample for evaluating textual descriptions. (Blue Jay class of CUB 200 2011 dataset)}
\vspace{-0.2cm}
\end{figure}

We hypothesize that the performance gains in cross-modal transfer training with proponent texts, compared to image-only training and other description-generating baselines, stem from the model’s ability to integrate complementary information from both modalities. Training with images and textual descriptions enables the model to jointly utilize visual features and contextual information. Unlike general textual descriptions from baselines, our proponent texts are filtered using influence estimation and CLIP scores, ensuring greater relevance and informativeness. Furthermore, by incorporating external knowledge, our descriptions help the model learn richer semantic representations of image classes.
The qualitative comparison in Figure \ref{fig:human_eval_main} demonstrates the differences in textual descriptions generated by various methods for the Blue Jay class of CUB 200 2011~\cite{wah2011caltech}. Menon~\cite{menon2022visual} provides the most basic descriptions, mentioning only the color of bird and a few physical traits, while LaBO~\cite{yang2023language} and CuPL~\cite{pratt2023does} offer more detailed explanations but with some repetition and a lack of additional visual features. VDT-Adapter~\cite{maniparambil2023enhancing} focuses on specific features like the eyestripe and beak but presents a more mechanical, segmented description. 
In contrast, our method delivers richer, more detailed descriptions, emphasizing unique facial features such as a 'short, robust beak, a crest on the head, and vibrant blue feathers,' providing a more complete and expressive visual image than the other methods.

\subsection{Ablation Studies} 
\label{sec:ablation}

\begin{table}[h]
\resizebox{\columnwidth}{!}{%
\begin{tabular}{c|c|c|c|c}
\hline
Dataset      & Only Images & \begin{tabular}[c]{@{}c@{}}IF\end{tabular} & \begin{tabular}[c]{@{}c@{}}CLIP\end{tabular} & \begin{tabular}[c]{@{}c@{}}IFT\end{tabular} \\ \hline
CUB 200 2011 & 71.332\%    & 72.834\% & 72.696\% & \textbf{74.525\%} \\
OxfordPets   & 91.664\%    & 92.722\% & 92.318\% & \textbf{93.396\%} \\
CIFAR-10     & 94.320\%    & 94.680\% & 94.730\% & \textbf{94.820\%} \\
CIFAR-100    & 77.830\%    & 77.910\% & 77.970\% & \textbf{78.250\%} \\
EuroSAT      & 94.296\%    & 94.222\% & 94.407\% & \textbf{96.037\%} \\
Food101      & 85.848\%    & 85.974\% & 85.934\% & \textbf{86.950\%} \\
Miniimagenet & 91.980\%    & 92.040\% & 92.020\% & \textbf{92.480\%} \\
102flowers   & 96.459\%    & 96.460\% & 96.337\% & \textbf{97.192\%} \\
DTD          & 72.074\%    & 73.989\% & 73.830\% & \textbf{74.043\%} \\ \hline
\end{tabular}%
}
\caption{
\label{t:ablation1}
Accuracies of ablation studies on Cross-Modal Transfer Classification with different proponent text determination methods: Influence score(IF), CLIP score(CLIP), and IFT.
}
\end{table}

\begin{figure*}
\centering
\includegraphics[width=1
\textwidth]{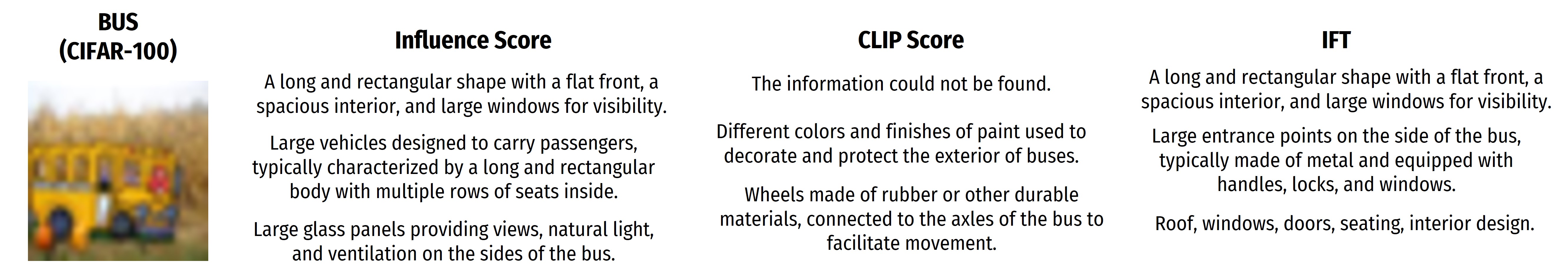}
\caption{Examples of selected proponent texts when using influence score (Influence Score), CLIP score (CLIP Score), and IFT (Bus class of CIFAR-100 dataset).}
\label{fig:ablation1}
\vspace{-0.2cm}
\end{figure*}

To verify the effectiveness of IFT as a scoring metric, we compare cross-modal transfer classification performance when using either the influence score or CLIP score alone instead of IFT in Table~\ref{t:ablation1}.
For all datasets, compared to when only the influence score or CLIP score is used, the performance is higher when cross-modal transfer training with proponent texts selected using IFT.

We also show examples of proponent texts that change when using only the influence score or CLIP score, and IFT in Figure \ref{fig:ablation1}. 
In this figure, one can see that proponent texts selected using only the CLIP score are just error messages, and rather about the decoration of the bus' appearance and the material of the bus' wheel. It is hard to say that this information is helpful as visual cues to classify the bus images. 
The proponent texts selected using the influence score contain information about the appearance of not only the bus but also the structure inside the bus. 
On the other hand, all of the proponent texts determined using IFT provide clues to classify the bus images, and one can see what factors help the model to make a decision. 
Through IFT, we can resolve issues mentioned in baseline methods, such as visual cues that do not helpful for models when classifying images, or the same text being repeated. Through this, we can see that IFT is appropriate as a score metric for proponent texts.

\begin{table}
\centering
\resizebox{0.9\columnwidth}{!}{%
\begin{tabular}{c|c|c|c}
\hline
Dataset      & Only Images & \begin{tabular}[c]{@{}c@{}}No Wiki\end{tabular} & \begin{tabular}[c]{@{}c@{}}Wiki\end{tabular} \\ \hline
CUB 200 2011 & 71.332\%    & 72.092\% & \textbf{74.525\%} \\
OxfordPets   & 91.664\%    & 92.453\% & \textbf{93.396\%} \\
CIFAR-10     & 94.320\%    & 94.490\% & \textbf{94.820\%} \\
CIFAR-100    & 77.830\%    & 77.410\% & \textbf{78.250\%} \\
EuroSAT      & 94.296\%    & 94.333\% & \textbf{96.037\%} \\
Food101      & 85.848\%    & 86.832\% & \textbf{86.950\%} \\
Miniimagenet & 91.980\%    & 91.570\% & \textbf{92.480\%} \\
102flowers   & 96.459\%    & 96.703\% & \textbf{97.192\%} \\
DTD          & 72.074\%    & 73.830\% & \textbf{74.043\%} \\ \hline
\end{tabular}%
}
\caption{
\label{t:ablation2}
Accuracies of ablation studies on Cross-Modal Transfer Classification with(Wiki) and without Wikipedia url provided to GPT-3.5(No Wiki).
}
\vspace{-0.5cm}
\end{table}

\begin{figure}
\centering
\includegraphics[width=8cm]{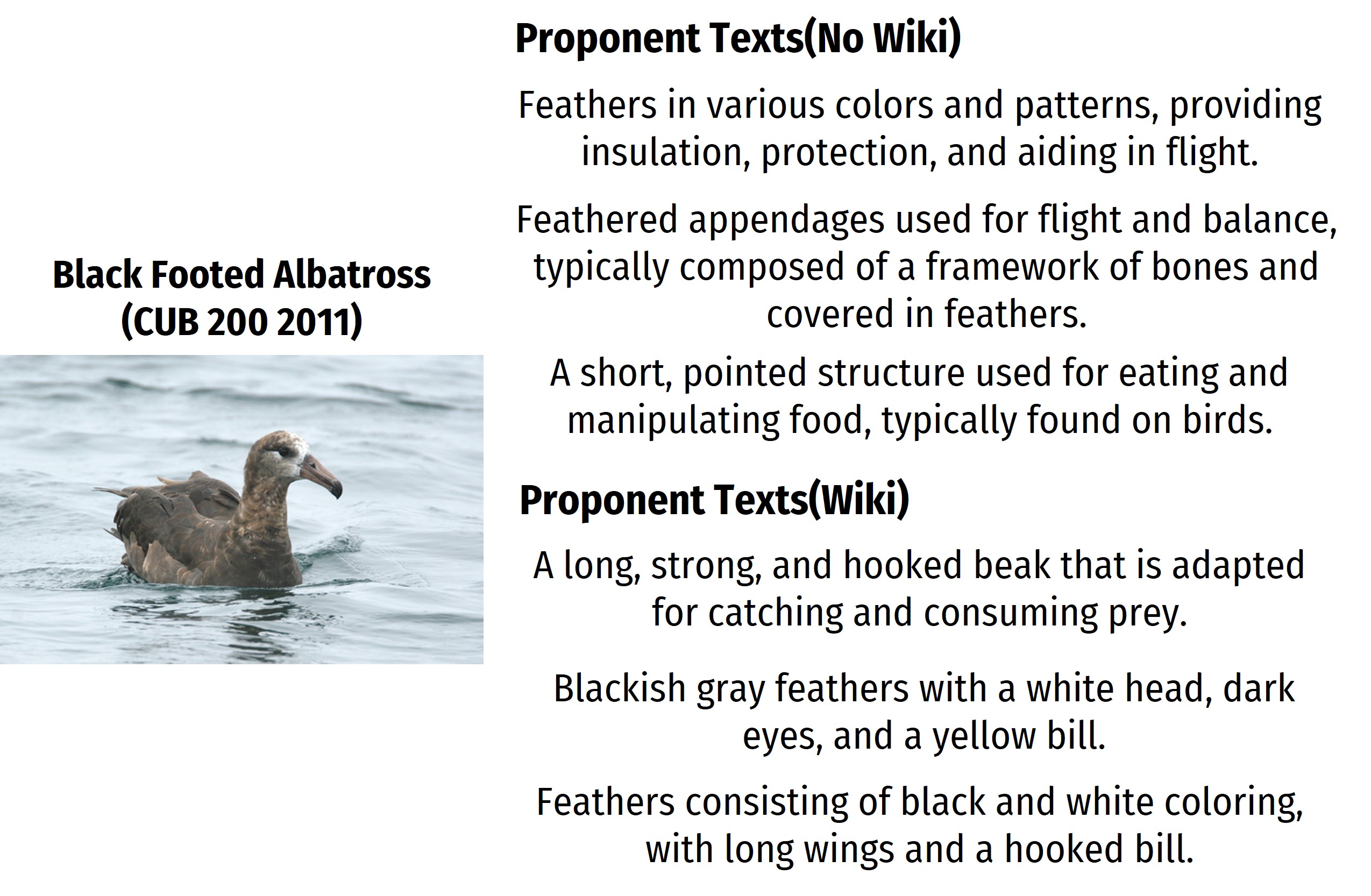}
\caption{Examples of the selected proponent texts when we provide Wikipedia urls to GPT-3.5 or not. (Black Footed Albatross class of CUB 200 2011 dataset)}
\label{fig:ablation2}
\vspace{-0.2cm}
\end{figure}

To evaluate the effectiveness of our 2-stage prompting with Wikipedia URLs, we compare cross-modal transfer performance when GPT-3.5 is provided with Wikipedia URLs versus when it is not. Table \ref{t:ablation2} presents the results, showing that across all datasets, providing Wikipedia URLs improves performance. Figure \ref{fig:ablation2} presents examples of proponent texts generated with Wikipedia URLs.
Without Wikipedia URLs, the selected proponent texts provide only general information about birds rather than class-specific details, such as "Feathered appendages used for flight and balance" or "Feathers in various colors and patterns, providing insulation, protection, and aiding in flight." In contrast, when Wikipedia URLs are used, the descriptions become more class-specific, such as "Blackish gray feathers with a white head, dark eyes, and a yellow bill" or "Feathers consisting of black and white coloring, with long wings and a hooked bill."
Additionally, utilizing external knowledge bases helps mitigate the hallucination effect of GPT-3.5. Figure \ref{fig:ablation2} shows that descriptions generated with external knowledge correctly depict the beak as long, whereas those generated without it inaccurately describe the beak as short. 
These results demonstrate that our 2-stage prompting with Wikipedia URLs effectively provides class-specific details while reducing hallucinations.

\subsection{Qualitative Evaluation with GPT-4o} \label{sec:third-party evaluation}
We additionally conduct evaluations using GPT-4o to compare the quality of textual descriptions generated by our baselines. For this experiment, we randomly select 100 classes from the nine benchmark datasets. From each class, we sampled three descriptions per method (CuPL, LaBo, Menon, VDT, and Ours), along with two original images as visual references. These two images and the five descriptions (one for each method) are then provided to GPT-4o for evaluation across three criteria: Helpful, Informative, and Relevant. This setup produces three evaluation instances per class, yielding 300 evaluation cases in total. We adopt two metrics: 1) \textbf{Top-1 Rating}~\cite{lin2023argue}: The proportion of times each method is rated as the best for a given criterion, 2) \textbf{Ranking Average}~\cite{xu2022sequence}: The mean rank of each method, where ties are permitted when descriptions are of similar quality. Table \ref{t:TP_top} reports the Top-1 Ratings, and Table \ref{t:TP_Ranking} presents the Ranking Averages. 

Our method consistently ranked highest across all metrics and criteria, demonstrating strong performance in generating helpful, informative, and relevant descriptions. The detailed evaluation process and prompt templates for GPT-4o evaluations are provided in Appendix~\ref{apd:thirdparty_detail}.

\begin{table}[]
\resizebox{\columnwidth}{!}{%
\begin{tabular}{c|l|l|l|l|l}
\hline
\multicolumn{1}{l|}{} & \multicolumn{1}{c|}{Menon} & \multicolumn{1}{c|}{LaBo} & \multicolumn{1}{c|}{CuPL} & \multicolumn{1}{c|}{VDT} & \multicolumn{1}{c}{Ours} \\ \hline
Helpful & 9.87\% & 8.88\% & 20.72\% & 29.28\% & \textbf{31.25\%} \\
Informative & 0.66\% & 15.13\% & 32.57\% & 21.71\% & \textbf{33.55\%} \\
Relevant & 7.89\% & 9.54\% & 27.63\% & \textbf{28.29\%} & \textbf{28.29\%} \\ \hline
\end{tabular}%
}
\caption{
\label{t:TP_top}
Top-1 Ratings (\%) from GPT-4o for evaluation of textual descriptions across three criteria: Helpful, Informative, and Relevant.
}
\end{table}

\begin{table}[]
\centering
\resizebox{0.9\columnwidth}{!}{%
\begin{tabular}{c|c|c|c|c|c}
\hline
 & Menon & LaBo & CuPL & VDT & Ours \\ \hline
Helpful & 2.31 & 1.99 & 2.11 & 2.02 & \textbf{1.6} \\
Informative & 2.61 & 2.09 & 1.90 & 1.95 & \textbf{1.57} \\
Relevant & 2.83 & 2.62 & 2.31 & 2.28 & \textbf{1.69} \\ \hline
\end{tabular}%
}
\caption{
\label{t:TP_Ranking}
Average ranks (lower is better) from GPT-4o for evaluation of textual descriptions across three criteria: Helpful, Informative, and Relevant.
}
\vspace{-0.2cm}
\end{table}

\section{Conclusion}
\label{sec:conc}

We propose a simple yet effective approach that generates textual descriptions that can explain the image data well using 2-stage prompting that utilizes external knowledge. 
By leveraging the generated textual description with the proposed IFT, defined as the sum of influence score and CLIP score, we can determine \emph{proponent texts}, which are informative texts for explaining each image class. 
Furthermore, we propose a novel benchmark task named \emph{cross-modal transfer classification}. By training the vision model with these proponent texts, we achieve improved performance compared to training with images alone.
Our approach enables a language-based understanding of data, potentially enhancing the interpretability of model predictions.
In quantitative evaluations, including zero-shot classification and cross-modal transfer classification, our method consistently outperforms baselines.
Ablation studies, qualitative analyses, and GPT-4o evaluation collectively validate that our generated descriptions are more helpful in model training than those of the baselines.


\section{Limitation}
\label{sec:limit}

This work has several limitations. First, the extracted textual descriptions can vary depending on the pre-defined prompts and the large language model(LLM) used, affecting performance. Second, as the dataset size increases, so does the computational load required to calculate influence scores, resulting in higher computational costs. In our future work, we plan to experiment with various prompts and large language models and also explore methods to efficiently compute influence scores even as dataset sizes increase.
Third, since our method relies on LLMs such as GPT-3.5/4, it may inherit the inherent biases present in these models. While we leverage external knowledge sources (e.g., Wikipedia URLs) to improve factual consistency and mitigate hallucinations, these strategies may not fully resolve deeper representational or social biases embedded in the LLMs' pretraining data.

\section*{Acknowledgments}
This work was supported by Institute of Information \& communications Technology Planning \& Evaluation (IITP)
grant funded by the Korea government (MSIT) (No.RS-2022-II220608/2022-0-00608, Artificial intelligence research about
multimodal interactions for empathetic conversations with humans, No.IITP-2025-RS-2024-00360227, Leading Generative AI Human Resources Development, No. RS-2025-25442824, AI Star Fellowship Program(Ulsan National Institute of Science and Technology, No. RS-2021-II212068, Artificial Intelligence Innovation Hub \& No.RS-2020-II201336, Artificial Intelligence graduate school support(UNIST)) and the
National Research Foundation of Korea(NRF) grant funded by the Korea government(MSIT) (No. RS-2023-00219959).

\bibliography{custom}

\appendix

\appendix
\section{2-stage Prompting with Wikipedia url}
\label{apd:prompting}

Since large language models generate different quality of answers depending on prompting, we provide more detailed implementation details of our 2-stage prompting with the Wikipedia url. We refer to the setting of baseline method~\citep{menon2022visual}. We use the prompting structure of Q: A: and provide GPT-3.5 desired output examples.

First, when we are extracting components of each class, we remind the form of questions:
\newline 
\newline 
\emph{Q : Can you tell me the components of $\{$class name$\}$ from the perspective of appearance?}
\newline    
\emph{A : }
\newline 

Prior to the above question, we provide additional examples like this:
\newline 
\newline 
\emph{Q : Can you tell me the components of American Bulldog from the perspective of appearance?}
\newline
\emph{
A :    1. Coat Type and Texture
	2. Coat Color
3. Body Build
4. Size
5. Head
6. Muzzle and Nose
7. Eyes
8. Ears
9. Tail
10. Legs and Paws
11. Coat Patterns
12. Facial Features
13. Unique Breed Traits
}
\newline 

\begin{table*}[]
\centering
\resizebox{0.9\textwidth}{!}{%
\begin{tabular}{c|c|c|c|c|c|c}
\hline
\textbf{Dataset} & \textbf{Zero-Shot} & \textbf{Menon} & \textbf{LaBo} & \textbf{CuPL} & \textbf{VDT-Adapter} & \textbf{Ours} \\ \hline
CUB 200 2011 & 62.703\% & 63.807\% & 63.928\% & \underline{65.084\%} & 64.273\% & \textbf{66.396\%} \\
OxfordPets & 89.488\% & 88.679\% & 88.814\% & \textbf{93.127\%} & 92.453\% & \underline{92.562\%} \\
CIFAR10 & \underline{93.660\%} & 92.640\% & 93.570\% & 93.120\% & 93.099\% & \textbf{93.819\%} \\
CIFAR100 & \underline{75.690\%} & 75.150\% & 74.880\% & 74.510\% & 75.250\% & \textbf{75.730\%} \\
EuroSAT & 40.778\% & 38.630\% & 41.815\% & \textbf{45.444\%} & 42.926\% & \underline{43.740\%} \\
Miniimagenet & 85.910\% & 85.590\% & 85.530\% & 86.220\% & \underline{86.280\%} & \textbf{86.435\%} \\
102flowers & 65.690\% & 68.987\% & 70.085\% & 70.452\% & \underline{71.062\%} & \textbf{72.161\%} \\
DTD & 51.968\% & 51.277\% & 54.628\% & \underline{58.404\%} & \underline{58.404\%} & \textbf{59.989\%} \\ \hline
\end{tabular}%
}
\caption{Zero-shot image classification results with laion/CLIP-ViT-L-14-laion2B~\cite{cherti2023reproducible} model.}
\label{tab:appendix_clip_new}
\end{table*}

Since we provide these examples to GPT-3.5, when we ask a question, the answers in the order of 1, 2, 3 are given. So we can get the desired answers by simply removing the numbers.

Second, when we are extracting textual descriptions of each class using the components of each class and the Wikipedia url, we remind the form of questions:
\newline
\newline
\emph{Q : Please summarize the information of appearance about $\{$components of each class$\}$ in this $\{$Wikipedia url$\}$ in one line composed of nouns. If you couldn't find related information, you must answer general information you know.}
\emph{A : } 
\newline

Prior to the above question, we also provide additional examples like this:
\newline
\newline
\emph{Q : Please summarize the information of appearance about nose in this url $https://en.wikipedia.org/wiki/American Bulldog$ in one line composed of nouns. If you couldn't find related information, you must answer general information you know like the above questions. }
\newline
\emph{
A : A short to medium-length muzzle with a nose that can be black, brown, or pigmented, often matching the coat color, and it is a distinctive feature on the breed's square-shaped head.
}
\newline

Since we ask for a "composed of nouns" answer, we can get a detailed and informative answer without refining it.

\section{Proponent Texts and IFT} 
\label{apd:prop_exp}

\begin{figure*}
\centering
\includegraphics[width=15cm]{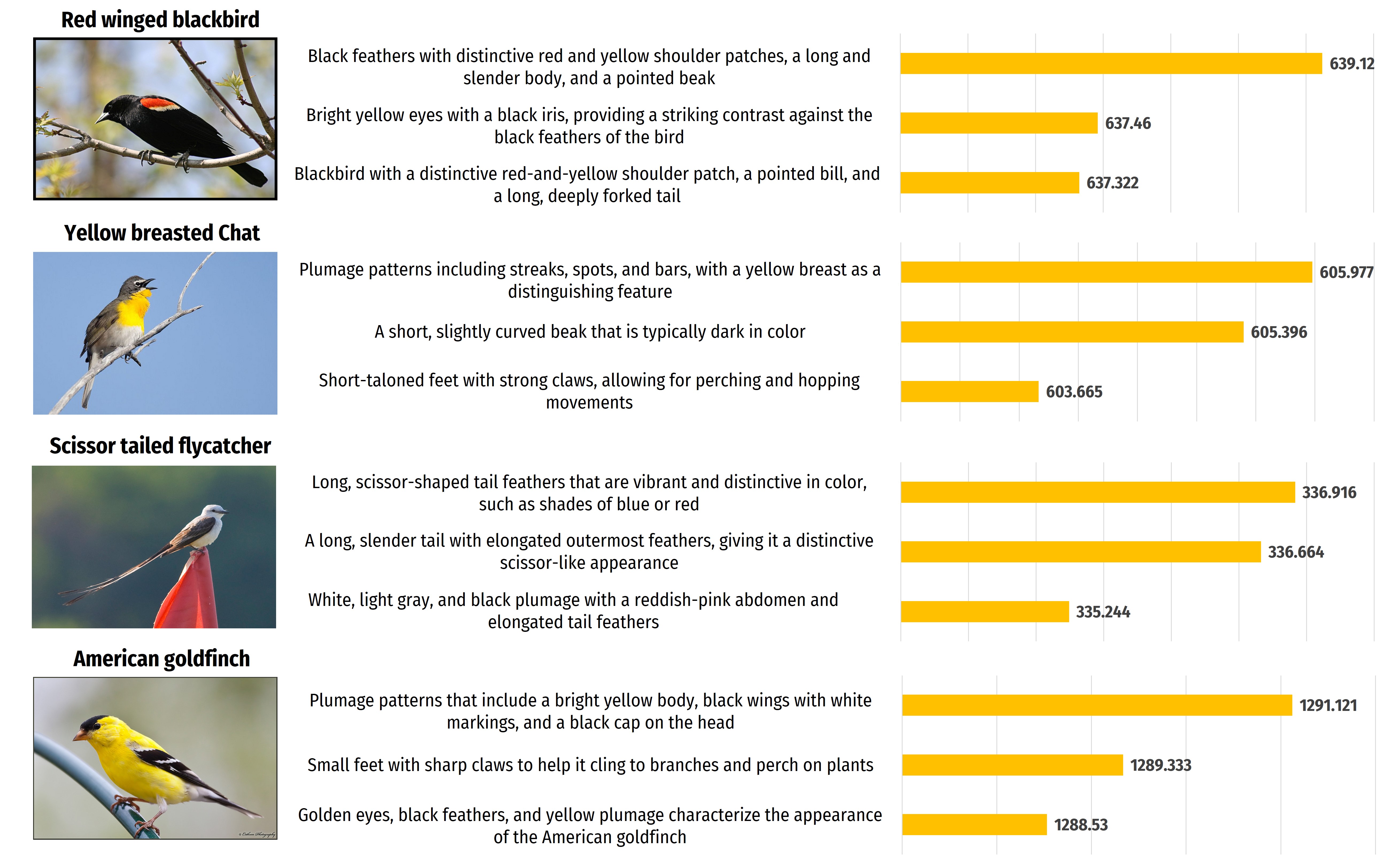}
\caption{Examples of proponent images, proponent texts and IFT for four classes of CUB 200 2011 dataset.}
\label{fig:apd_prop_exp1}
\end{figure*}

\begin{figure*}
\centering
\includegraphics[width=15cm]{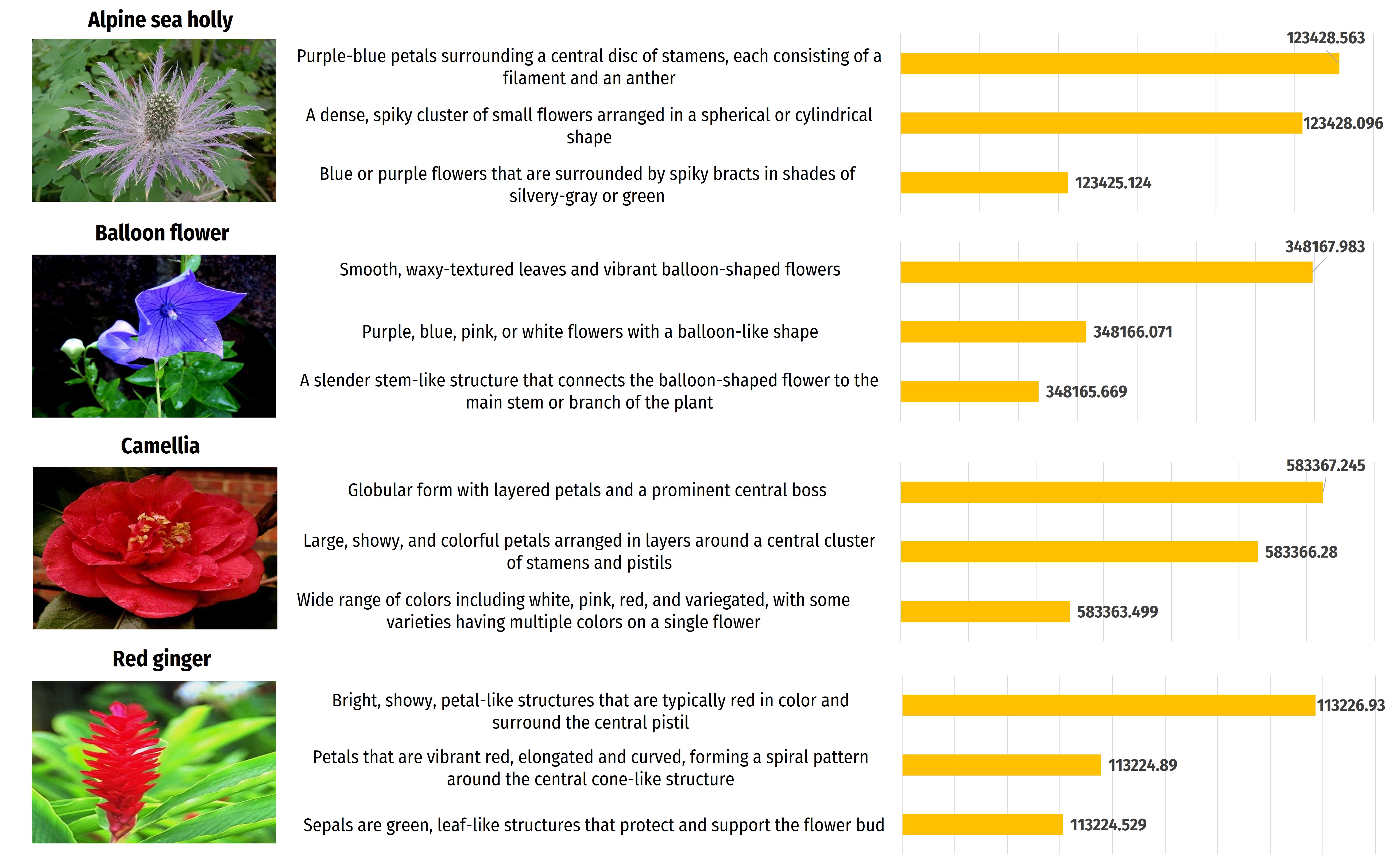}
\caption{Examples of proponent images, proponent texts and IFT for four classes of 102flowers dataset.}
\label{fig:apd_prop_exp2}
\end{figure*}

\begin{figure*}
\centering
\includegraphics[width=15cm]{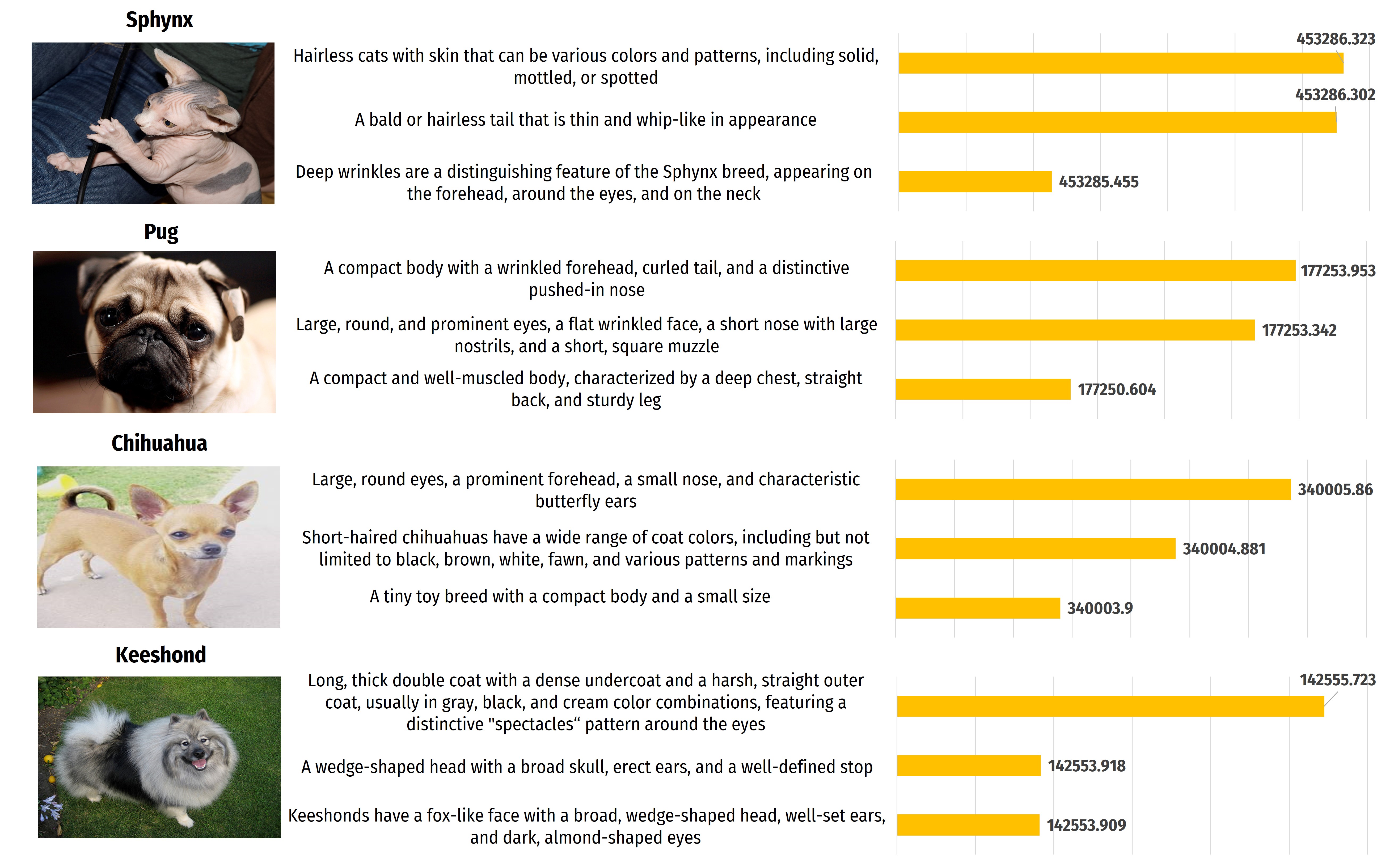}
\caption{Examples of proponent images, proponent texts and IFT for four classes of OxfordPets dataset.}
\label{fig:apd_prop_exp3}
\end{figure*}

\begin{figure*}
\centering
\includegraphics[width=15cm]{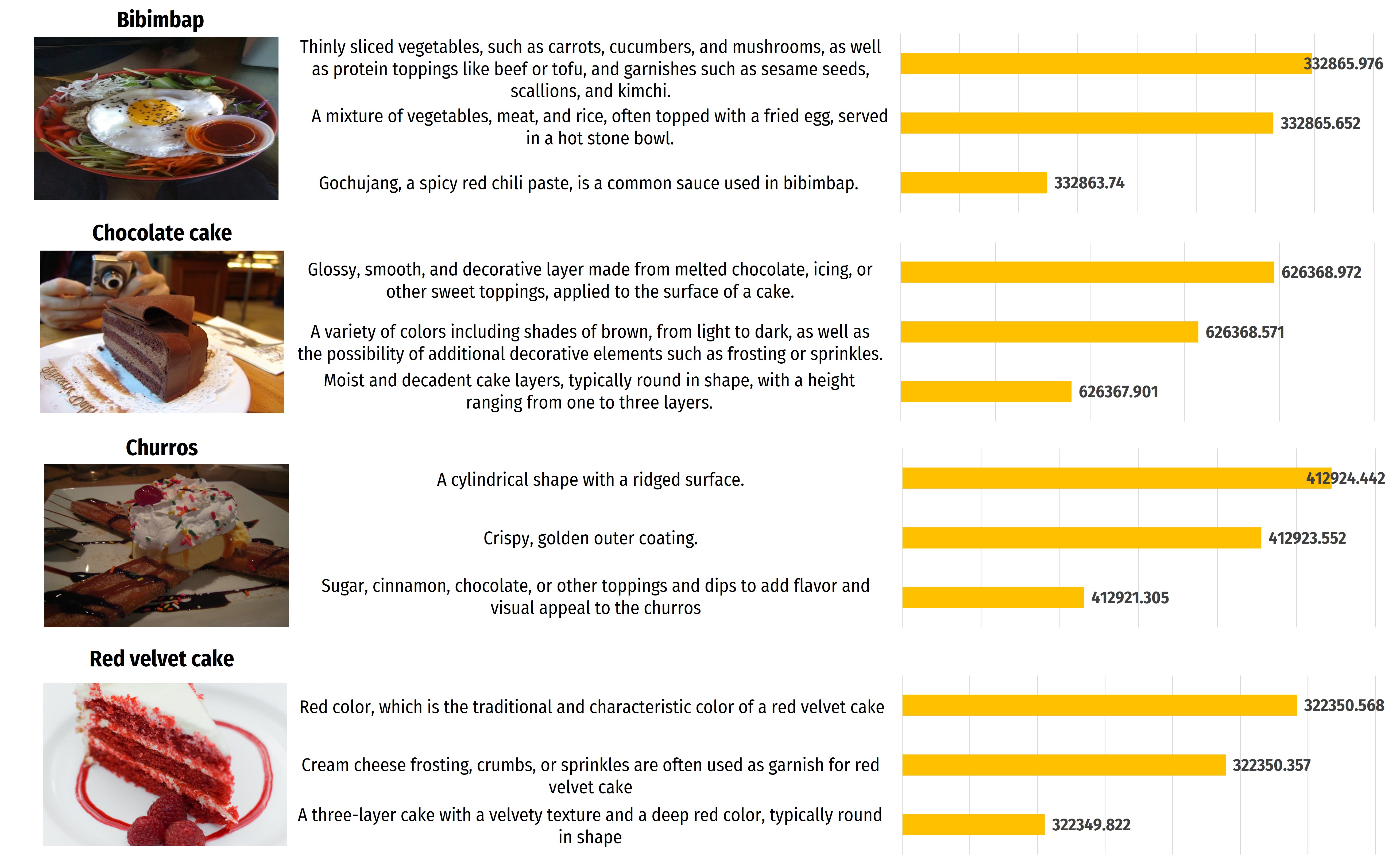}
\caption{Examples of proponent images, proponent texts and IFT for four classes of Food101 dataset.}
\label{fig:apd_prop_exp4}
\end{figure*}

\begin{figure*}
\centering
\includegraphics[width=15cm]{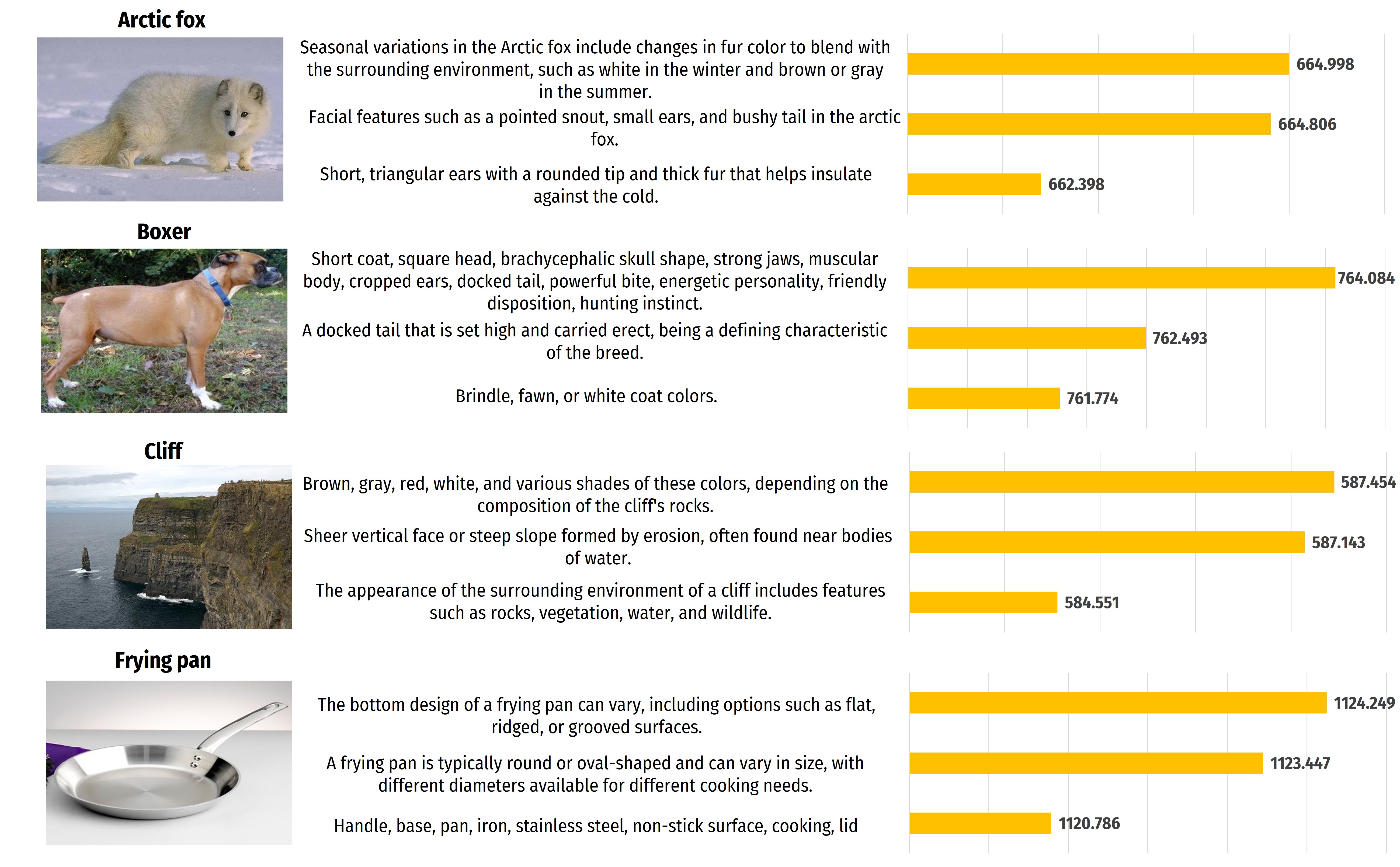}
\caption{Examples of proponent images, proponent texts and IFT for several classes of Miniimagenet dataset.}
\label{fig:apd_prop_exp5}
\end{figure*}

In this section, we provide additional examples of proponent texts and IFT in Figure \ref{fig:apd_prop_exp1}, \ref{fig:apd_prop_exp2}, \ref{fig:apd_prop_exp3}, \ref{fig:apd_prop_exp4}, \ref{fig:apd_prop_exp5}.

\begin{figure*}[h]
\centering
\includegraphics[width=0.88\textwidth]{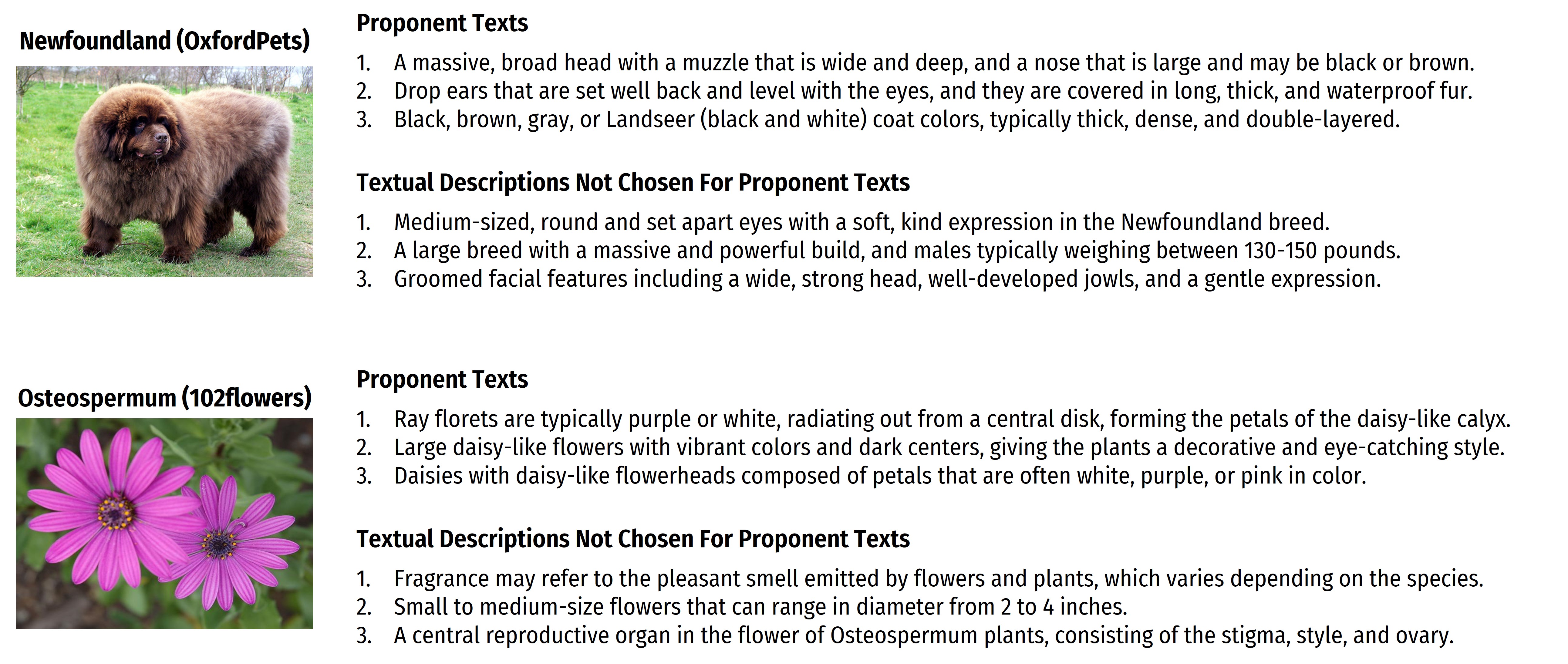}
\caption{Examples of proponent texts and non-proponent texts for the Newfoundland class of the OxfordPets dataset and the Osteospermum class of the 102flowers dataset.}
\label{fig:analysis}
\end{figure*}



\section{GPT-4o Evaluation Details}
\label{apd:thirdparty_detail}
To objectively assess the quality of textual descriptions generated by each method, we conduct a third-party evaluation using GPT-4o, a vision-language model with strong reasoning capabilities.

\paragraph{Evaluation Setup.}
We randomly select 100 classes across all nine benchmark datasets. For each class, we sample three descriptions from each of the five methods (Menon et al., LaBo, CuPL, VDT-Adapter, and Ours), resulting in 15 descriptions per class. We also sample two reference images per class from the original dataset. The evaluation is conducted in the zero-shot setting and produces 300 evaluation instances in total (3 per class).

\paragraph{Evaluation Criteria.}
For each evaluation instance, GPT-4o is presented with two reference images from the same class and five textual descriptions (one per method) in randomized order.
The model is then asked to evaluate each description according to the following three criteria:
\textbf{Helpfulness:} how useful the description is in understanding the visual content. \textbf{Informativeness:} how much specific and relevant detail the description provides. \textbf{Relevance:} how well the description aligns with the given images.

\paragraph{Prompt Template for Top-1 Rating.}
The following prompt is used to collect Top-1 rating judgments from GPT-4o. The model is instructed to return only the highest-ranked description for each evaluation criterion, along with a brief reasoning.

\emph{You are a vision-language model evaluator.}

\emph{Given two images and five textual descriptions, your task is to rank the descriptions for each of the following three criteria and output only the top-1 ranking description (or group of descriptions if equally best) along with a short rationale. The criteria are:}

\emph{1. \textbf{Helpful}: Does the description help distinguish or understand the two images effectively?}

\emph{2. \textbf{Informative}: Does the description provide detailed and meaningful content?}

\emph{3. \textbf{Relevant}: Does the description accurately reflect the visual content of the two images?}

\emph{You are allowed to assign the same rank to multiple descriptions if you believe they are equally strong for a given criterion, but please output only the top-ranked group for each criterion.}

\emph{Images:}

\emph{Image A: <Insert Image A>}

\emph{Image B: <Insert Image B>}

\emph{Descriptions:}

\emph{1. "\{description1\}"}

\emph{2. "\{description2\}"}

\emph{3. "\{description3\}"}

\emph{4. "\{description4\}"}

\emph{5. "\{description5\}"}

\emph{For each criterion, please output only the top-1 ranking description(s) along with a short rationale for why these descriptions are the best.}

\emph{Output format:}

\emph{\#\#\# Helpful Ranking:}

\emph{\textbf{Top-1}: Descriptions 1}

\emph{Reason: "These descriptions clearly highlight key differences between the two images, such as beak shape and feather patterns."}

\emph{---}

\emph{\#\#\# Informative Ranking:}

\emph{\textbf{Top-1}: Descriptions 2}

\emph{Reason: "They include specific visual details such as color, size, and structural features."}

\emph{---}

\emph{\#\#\# Relevant Ranking:}

\emph{\textbf{Top-1}: Description 5}

\emph{Reason: "Highly aligned with actual visible features in both images."}

\emph{Please ensure your ranking is thoughtful and grounded in what is visible in the two images.}

\paragraph{Prompt Template for Mean Ranking.}
The following prompt is used to collect mean ranking scores from GPT-4o for each evaluation instance. The model is instructed to rank five descriptions per criterion, with ties allowed.

\emph{You are a vision-language model evaluator.}

\emph{Given two images and five textual descriptions, your task is to rank the descriptions for each of the following three criteria:}

\emph{1. \textbf{Helpful}: Does the description help distinguish or understand the two images effectively?}

\emph{2. \textbf{Informative}: Does the description provide detailed and meaningful content?}

\emph{3. \textbf{Relevant}: Does the description accurately reflect the visual content of the two images?}

\emph{You are allowed to assign the same rank to multiple descriptions if you believe they are equally strong for a given criterion.}

\emph{Images:}

\emph{Image A: <Insert Image A>}

\emph{Image B: <Insert Image B>}

\emph{Descriptions:}

\emph{1. "\{description1\}"}

\emph{2. "\{description2\}"}

\emph{3. "\{description3\}"}

\emph{4. "\{description4\}"}

\emph{5. "\{description5\}"}

\emph{For each criterion, please list the descriptions grouped by rank, with a short rationale for each group.}

\emph{Output format:}

\emph{\#\#\# Helpful Ranking:}

\emph{\textbf{Rank 1}: Descriptions 2, 5}

\emph{Reason: "These descriptions clearly highlight key differences between the two images, such as beak shape and feather patterns."}

\emph{\textbf{Rank 2}: Description 1}

\emph{Reason: "Provides some useful context but lacks comparative elements."}

\emph{\textbf{Rank 3}: Descriptions 3, 4}

\emph{Reason: "These are vague or unrelated to distinguishing the images."}

\emph{---}

\emph{\#\#\# Informative Ranking:}

\emph{\textbf{Rank 1}: Descriptions 1, 5}

\emph{Reason: "They include specific visual details such as color, size, and structural features."}

\emph{...}

\emph{\#\#\# Relevant Ranking:}

\emph{\textbf{Rank 1}: Description 5}

\emph{Reason: "Highly aligned with actual visible features in both images."}

\emph{...}

\emph{Please make sure your ranking is thoughtful and grounded in what is visible in the two images.}

\begin{figure*}
\centering
\includegraphics[width=1\textwidth]{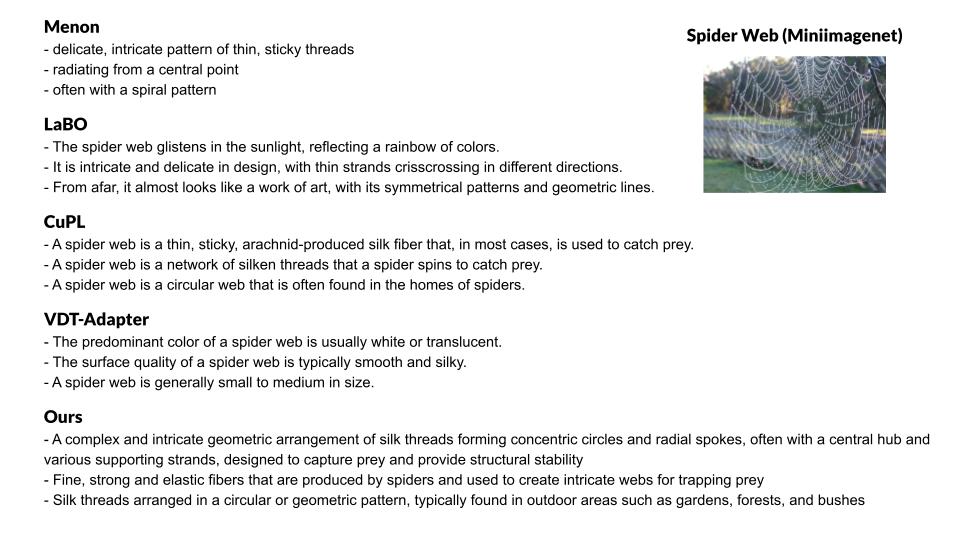}
\caption{\label{fig:apd_quali1} Example of a qualitative sample for evaluating textual descriptions. (Spider Web class of Miniimagenet dataset)}
\end{figure*}

\begin{figure*}
\centering
\includegraphics[width=1\textwidth]{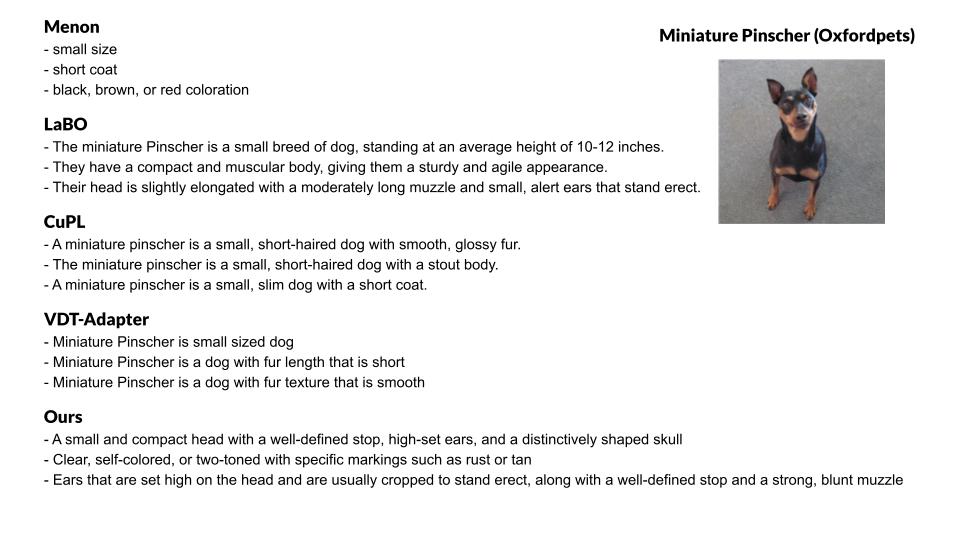}
\caption{\label{fig:apd_quali2} Example of a qualitative sample for evaluating textual descriptions. (Miniature Pinscher class of Oxfordpets dataset)}
\end{figure*}

\section{Analysis} 
\label{apd:analysis}

In this section, we demonstrate that IFT can select texts that describe each image class well by showing examples of proponent texts and non-proponent texts.

Figure \ref{fig:analysis} shows examples of texts selected as proponent texts and those that are not.
For the Newfoundland class of the OxfordPets dataset, non-proponent texts contain general information such as expression or weight, which do not aid in classifying Newfoundland images. For the Osteospermum class of the 102flowers dataset, non-proponent texts include information about fragrance, growth, and reproductive organs, which are not helpful for image classification.

These examples illustrate that IFT ensures proponent texts contain information that effectively helps the model classify images. 

\section{Dataset Details} 
\label{apd:dataset}

\begin{table}[h]
\centering
\resizebox{\linewidth}{!}{
\begin{tabular}{c|c|c|c|c}
\hline
Dataset & Classes & Train Size & Validation Size & Test Size \\ \hline
CUB 200 2011 & 200 & 4800 & 1194 & 5794 \\
Miniimagenet & 100 & 40000 & 10000 & 10000 \\
CIFAR-10 & 10 & 40000 & 10000 & 10000 \\
CIFAR-100 & 100 & 40000 & 40000 & 10000  \\
OxfordPets & 37 & 5910 & 697 & 742 \\
Food101 & 101 & 75750 & 10100 & 15150\\
EuroSAT & 10 & 18900 & 5400 & 2700 \\ 
102flowers & 102 & 6552 & 818 & 819 \\
DTD & 47 &  1880& 1880& 1880\\ \hline
\end{tabular}
}
\caption{
Dataset partitions of the train, validation , and test set of total nine datasets we use.
}
\label{t:dataset}
\end{table}

We use a total of nine datasets. Additionally, we report the size of the train set, validation and test set of the datasets we use in Table \ref{t:dataset}. 

\textbf{CUB 200 2001~\citep{wah2011caltech}} This dataset contains a total of 200 bird species, with each species having around 30 images.  

\textbf{Miniimagenet~\citep{vinyals2016matching}} This dataset consists of 100 classes, each containing 600 images. The classes are drawn from a larger dataset called ImageNet~\citep{deng2009imagenet}, which contains over a million images across thousands of classes. It is introduced in~\citet{vinyals2016matching} as a benchmark dataset for few-shot learning, but we use 500 images for each class as a train set and 100 images as a test set. 

\textbf{CIFAR-10~\citep{krizhevsky2009learning}} This is a widely-used dataset in machine learning research, consisting of 60,000 32x32 color images in 10 classes. 

\textbf{CIFAR-100~\citep{krizhevsky2009learning}} This is an extension of CIFAR-10 dataset, consisting of 60,000 32x32 color images in 100 classes. 

\textbf{OxfordPets~\citep{parkhi2012cats}} The dataset is a collection of approximately 7,349 images containing 37 different breeds of cats and dogs. It is commonly used for tasks such as image classification and object detection in computer vision research.

\textbf{EuroSAT~\citep{helber2019eurosat}} This dataset is used for classifying geographical land cover types based on satellite imagery. It comprises 10 distinct geographical landscape classes, including categories like forests, cropland, roads, buildings, and rivers, among others. 

\textbf{Food101~\citep{bossard2014food}} The dataset is a widely used collection of food images that is primarily employed for food recognition and image classification tasks. It consists of approximately 101,000 images, each depicting a specific food item from one of the 101 distinct food categories.

\textbf{102flowers~\citep{nilsback2008automated}} The dataset is a collection of flower images, consisting of 102 different categories, each representing a distinct species or type of flower. 

\textbf{Describable Textures Dataset (DTD)~\citep{cimpoi2014describing}} This dataset is a collection of diverse texture images, containing 47 different categories with distinct visual features and structures.




\end{document}